\documentclass[letterpaper, 10 pt, conference]{ieeeconf}  

\IEEEoverridecommandlockouts                              

\usepackage{graphics} 
\usepackage{graphicx}
\usepackage{lscape}
\usepackage{booktabs}
\usepackage{pgfplots}
\usepackage{pgfplotstable}

\title{\LARGE \bf
A Survey of Robotics and Emotion: \\ 
Classifications and Models of Emotional Interaction
}

\author{Richard Savery and Gil Weinberg $^{1}$ 
\thanks{$^{1}$Georgia Tech Center for Music Technology}
\thanks{\tt\small rsavery3@gatech.edu, gilw@gatech.edu}}%

\begin{document}

\maketitle
\thispagestyle{empty}
\pagestyle{empty}

\begin{abstract}
As emotion plays a growing role in robotic research it is crucial to develop methods to analyze and compare among the wide range of approaches. To this end we present a survey of 1427 IEEE and ACM publications that include robotics and emotion. This includes broad categorizations of trends in emotion input analysis, robot emotional expression, studies of emotional interaction and models for internal processing. We then focus on 232 papers that present internal processing of emotion, such as using a human’s emotion for better interaction or turning environmental stimuli into an emotional drive for robotic path planning. We conducted constant comparison analysis of the 232 papers and arrived at three broad categorization metrics - emotional intelligence, emotional model and implementation - each including two or three subcategories. The subcategories address the algorithm used, emotional mapping, history, the emotional model, emotional categories, the role of emotion, the purpose of emotion and the platform. Our results show a diverse field of study, largely divided by the role of emotion in the system, either for improved interaction, or improved robotic performance. We also present multiple future opportunities for research and describe intrinsic challenges common in all publications. 

\end{abstract}

\section{Introduction}
\begin{figure*}[h]
  \centering
\includegraphics[width=16cm]{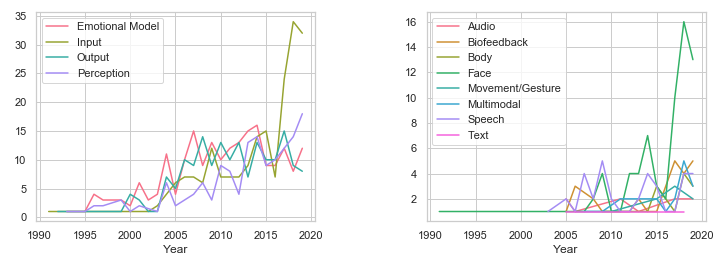}
  \caption{Emotional Categorizations from Abstracts and Sub-Categories from Inputs}
  \label{fig:broad}
\end{figure*}

Research in robotics and emotions has seen dramatic increases over the last thirty years (see Figure \ref{fig:broad}). This research has taken many forms, such as analyzing a person's emotional expression \cite{7783498}, or presenting believable robotic emotional output \cite{fukuda2004facial}. Publications have also focused on the role of emotion in human-robot interaction, and models for how robots should internally process and respond to emotion. This wide range of approaches and methodologies is not easily classified, analyzed or compared between projects. Currently when developing new research using robotics and emotion there is no standard practice or framework to easily place new work, forcing roboticists to continually make new choices drawing from psychology literature.

In this paper we conduct an extensive survey and meta-analysis on publications that combine robotics and emotion through any means. We begin by presenting broad categorizations of the types of inputs and outputs used by these systems. Our focus is then placed on systems that use emotion as part of their internal processing. This internal processing can be deeply varied, addressing aspects such as how robots can process and utilize information about humans' emotions, or how they can turn external stimuli into an emotional response that can then improve system performance. 

Our meta-analysis is based on collecting all publications from the IEEE XPlore digital library\footnote[2]{https://ieeexplore.ieee.org/Xplore/home.jsp} and the ACM Full-Text Collection\footnote[3]{https://dl.acm.org/} that discuss emotion and robotics. While IEEE and ACM do not include all publications on robotics and emotion, combined they contain 4 out of 5 of the robotics conferences and journals with the highest h-index. We believe that in combination, IEEE and ACM contain a representative and broad enough range of publications to develop an analysis of robotic research as a whole. Our search resulted in 1427 publications of which the abstracts were analyzed and classified. A constant comparison analysis was conducted on 232 papers that included emotional models. After establishing related works and our motivation this paper transitions to the Method section, which describes our literature review process and the categories that emerged through the analysis. This is followed by our results which presents an objective analysis of the data collected. We conclude with a discussion section presenting our insights and broad ideas for future work from the survey. Due to the quantity of publications analyzed, our reference section does not contain them all. Instead we  only cite works that are specifically mentioned throughout the paper. A full list of analyzed publications is available online. \footnote{https://github.com/richardsavery/robot-emotions-survey}

\section{Background and Motivation}
Emotions are a widely studied phenomena with many classification methods. The most prominent discrete categorization is that proposed by Ekman \cite{ekman1999basic} and includes fear, anger, disgust, sadness, happiness and surprise. Another way to consider emotions is with a continuous scale, most common is the Circumplex model; a two dimension model using valence and arousal \cite{posner2005circumplex}. Mood is often considered a longer term form of emotion, taking place over longer spans than emotion \cite{watson1994emotions}, while affect is human experience of feeling emotions. For the purpose of this analysis we consider the term emotion in the broadest sense, and include analysis of papers whether they focus on emotion, affect or mood. Similarly, for the term robot we include all publications that describe a plan or potential to implement in robotics in the future, or are published in robotic conferences.

Research into robotics and emotion can be divided into two main categories, emotion for social interaction, and emotion for improved performance and ``survivability'' \cite{arkin2009ethical}. For interaction, emotion can be used to improve agent likeability and believability \cite{ogata2000emotional}. Emotion in interaction has also been used to improve communication and allow for intuitive dialogue between human and robot \cite{breazeal2003emotion}. The second main purpose for implementing emotion in robotics is for improved performance or survivabilty. This builds on the belief that emotion is key to animals' ability to survive and navigate in the world and can likewise be applied to robotics \cite{arkin2003ethological}. 

There have been multiple surveys and meta-analysis on robots and human-robot interaction, although to our knowledge none focusing on emotion. In 2008 an extensive survey was conducted on human-robot interaction \cite{goodrich2008human} with only limited mention of emotion. Many other robot surveys have focused on specific aspects of robotics, such as robotic grasp \cite{shimoga1996robot}, social robotics \cite{leite2013social}, or empathy \cite{paiva2017empathy}. The closest publication addressing a survey on robotics and emotions is ``A Survey of Socially Interactive Robots''  \cite{fong2003survey} however, was written in 2002 and only contains a brief overview of emotion and robotics. Considering the rapid growth and interest in emotion and robotics, we believe that a meta-analysis of emotion and robotics with an emphasis on emotional interaction and modelling is now due.

\section{Method}
\begin{figure}[h]
  \centering
\includegraphics[width=8cm]{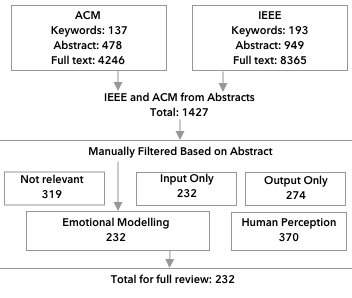}
  \caption{Flow Chart of Survey Method}
  \label{fig:method}
\end{figure}

Our review process was divided into three main steps, as done by Frich et al. \cite{frich2018twenty}. The first step involved finding all relevant articles and collecting publications. This was followed by dividing these papers into broad categories with a preliminary analysis. From there we conducted a thorough analysis on the remaining articles (see Figure \ref{fig:method}).

\subsection{Step 1: Collecting Publications}
We originally collected publications by retrieving all papers that contained the keywords Robot and Emotion from the IEEE Digital Library and the ACM Digital Collection. This resulted in a collection of 330 publications. From a random sampling of abstracts that contained the word emotion, without the keyword emotion we quickly realized that relying on keywords would not provide an extensive survey. We then expanded to collecting all publications that include the words robot and either emotion (or a variation such as emotional) or affect or mood. This resulted in a collection of 1,427 publications referencing robotics and emotion ranging from 1986 to February 2020.

\subsection{Step 2: Preliminary Sorting}
From the 1,427 publications all abstracts were manually read by the first author and in cases where the abstract was not clear, relevant sections of the paper itself were checked. We then sorted the articles into four categories. These categories were input focus, output focus, emotional modelling, and perception. We also created a separate list of articles that were not relevant for the survey. Figure \ref{fig:method} shows the quantity of each article from each collection, including duplicates between IEEE and ACM.

Our primary category focused on models of emotion and how robots can interact emotionally. These included a wide variety of systems discussed in detail in Section \ref{method:step3} and comprise the papers used in later sections. Input only publications focused on a method of input to a robotic system such as facial recognition \cite{mohseni2014facial} or speech recognition\cite{zhu2019emotion}. Output only papers focused on robots conveying some emotion and often evaluated the output, such as audio \cite{savery2019establishing} or robotic gait \cite{destephe2013conveying}. If the system included an emotional input and output it was placed in our primary category of emotional modelling. 

The category human perception included publications that discussed and evaluated existing robots perception to a range of audiences. These were occasionally ``wizard of oz'' setups with no clear path to implementation \cite{mok2014empathy}, or studies on interaction design that do not present new technologies \cite{lee2013aesthetic}, or general surveys of audience emotional attitudes to robots \cite{karim2016older}.

The list of publications that were deemed as irrelevant included a peripheral use of the word emotion, or measuring a users emotion when interacting with a robot for evaluation without the robot processing the emotion \cite{westlund2016transparency}. There were also occasional duplicates in the data-set and some extended-abstracts lacking sufficient detail to be categorized.

After placing each publication in categories we considered reducing the publications for analysis by citation count or citation average by year. We chose not to use this approach as we aimed to include as broad a range of approaches as possible, including experimental approaches and the full spectrum of divergent emotional modelling techniques. With this in mind we decided against removing potential approaches and concepts based solely on a paper's lack of visibility in citation counts.

\subsection{Step 3: Emotional Coding}\label{method:step3}
\begin{figure*}[h]
  \centering
\includegraphics[width=15cm]{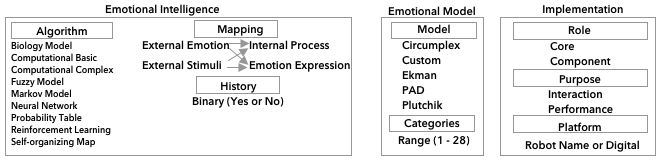}
  \caption{Categories Developed with Constant Comparison Analysis}
  \label{fig:categories}
\end{figure*}

Our final review method is based on the broad principles described by Onwuegbuzie and Frels \cite{onwuegbuzie2016seven}. We used constant comparison analysis to build categories and subcategories from the analyzed papers by coding each paper, then organizing codes and continual refinement of categorization. Through this process we developed three primary categories - emotional intelligence, emotional model and implementation (see Fig. \ref{fig:categories}). Emotional intelligence refers to how the system processes emotion, focusing on how input is translated through an algorithm to an output, and whether or not it contains a knowledge of past events or history. Emotional model analyzes what types of emotion are used and how many are used, while the final category, implementation categorizes the role and purpose of emotion and the platform that is used.

\subsubsection{Emotional Intelligence - Algorithm}
Our first categorization classifies the algorithm used to drive the system. There was a large range of approaches which did not easily break into categories, however multiple overarching trends did emerge. We found several reoccurring algorithms such as Fuzzy Models, Markov Models, Neural Networks, Probability Tables, Reinforcement Learning and Self-Organizing Maps. Each of these implementations varied greatly in complexity. For example Markov models could range from simple first order implementations \cite{qing2011artificial} to complex hidden Markov models \cite{li2018emotional}. 

In addition to these subcategories we found three other broader categories - biology inspired systems, computational models basic, and computational models complex. Biology inspired systems draw directly from comparisons to human or animal systems such as imitating a homeostasis approach \cite{malfaz2011biologically} or a neurocognitive affective system \cite{park2006neurocognitive}. Computational basic included simple implementations, which  used direct mappings, such as when something goes wrong being sad, or excited when asked for help \cite{antona2019my} or systems with clear tables of responses for different states \cite{diaz2018intelligent}. Computational complex featured custom systems that did not fit in the other categories and had more detailed models of emotion, including complicated mappings between all inputs \cite{canamero2001show}. Importantly we imply no superiority between basic and complex, as this only represents one component of the system and a more complex emotional model did not necessarily lead to better results. 

\subsubsection{Emotional Intelligence - History}
The category history referred to whether or not the model remembered or altered emotions based on past information. This category was binary, with the system either having history or not. A basic form of history involved a system that moves in a certain direction based on the input, such as moving in incremental steps between their past emotion and a human's current emotion \cite{shih2017implement}. In publications that featured longer history the term mood was often used, with combined emotion and mood models occurring very frequently \cite{gockley2006modeling,masuyama2014affective}.

\subsubsection{Emotional Intelligence - Mapping}
The category mapping described what input and output was used by the system. We developed two categories for input types and two for output types. These categories could work in any combination, such as both inputs to one output. The input categories were external stimuli and external emotion while the output categories were internal process and emotion expression. External stimuli include all stimuli that do not contain emotional information such as the distance from a wall or other perceived features \cite{tang2012robot}. External stimuli also include goals and tasks of a system, such as a robot's list of objectives \cite{izumi2009behavior}. External emotion primarily contains recognition of a human's emotions, such as through voice input \cite{lim2011converting}, but can also contain content that has been preassigned an emotion externally before use with the robot, such as emotionally tagged images \cite{diaz2018intelligent}. 
Emotional expression as an output occurs anytime the robot expresses emotion, such as through facial expressions \cite{hara1996real}. Internal process is when the robot uses an emotion internally to change or alter its decisions in a way that does not lead to an emotional expression. This form of output is common for environmental navigation and path planning \cite{lee2009mobile}.


\subsubsection{Emotional Model - Model and Categories}
Our aim was to present patterns of emotional models occurrences in publications related to robotics. With this goal in mind, emotional models were added and classified based on their presence in the publications analyzed, and not on their existence in emotional literature. Our final categorization included standard emotional models, Cirumplex (Valence/Arousal), Ekman's six categorizations, Plutchik's Wheel of Emotions \cite{plutchik2001nature} and PAD (3 dimensional model) \cite{mehrabian1980basic}. 

In addition to these categories we included a broad category for custom definitions. Custom models ranged from subjective variations of Ekman's categorizations, to original approaches, such as happy, hungry and tired \cite{ho1997model} that do not reference any other literature. These custom choices were often tailored to fit the robotic task at hand such as frustration \cite{murphy2002emotion} or combinations such as tired, tension and happiness \cite{li2008cooperative}. 

For each categorization we also included the amount of labelled emotions that were used. This number varied widely and often didn't allude to the variety of emotions for each model. For example, binary classifications were commonly not only happy and sad, or positive and negative, but could instead be courage and fear \cite{dominguez2006emotional}. When custom classifications used a single emotion this could be any emotion, such as guilt in a military application \cite{arkin2009ethical} or regret for optimal task queuing \cite{jiang2019respect}.

This classification specifically categorized the emotion used in the implemented model. Many papers referenced the Plutchik, Ekman, or the Circumplex model but made significant custom changes to the model. For clarity we always referenced the minimum amount of emotions used in a system. For example, if a system was able to detect 8 states from a face, but after processing displayed 4 different emotions we would count this as 4. This was however rare, publications almost always maintained the same model and emotion types throughout a system.

\subsubsection{Implementation - Role, Purpose and Platform}
The implementation category contained three subcategories - the role, the purpose and the robotic platform used. The role of emotion was either core or component, with core representing publications where the emotion was the central part of the implementation, and component when it is a part of a broader system. The classification labelled purpose analyzes how emotion acts within the system as a whole. Through the survey process, we arrived at two labels, interaction and performance. These two categories match those proposed by Arkin, who describes the purpose of emotions in robotics as interaction and 'survivability', which allows the robot to better interact with the world \cite{arkin2003moving}. The robotic platform considered the robot used for implementation, ranging from common HRI robots such as NAO \cite{nanty2013fuzzy}, to custom designs \cite{lee2008evolutionary} or digital interfaces \cite{nasir2018markov}.

\begin{figure}[b]
  \centering
\includegraphics[width=8cm]{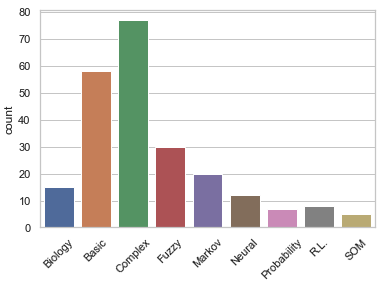}
  \caption{Algorithm Use Count}
  \label{fig:alogrithmusecoun}
\end{figure}

\begin{figure}[b]
  \centering
\includegraphics[width=8cm]{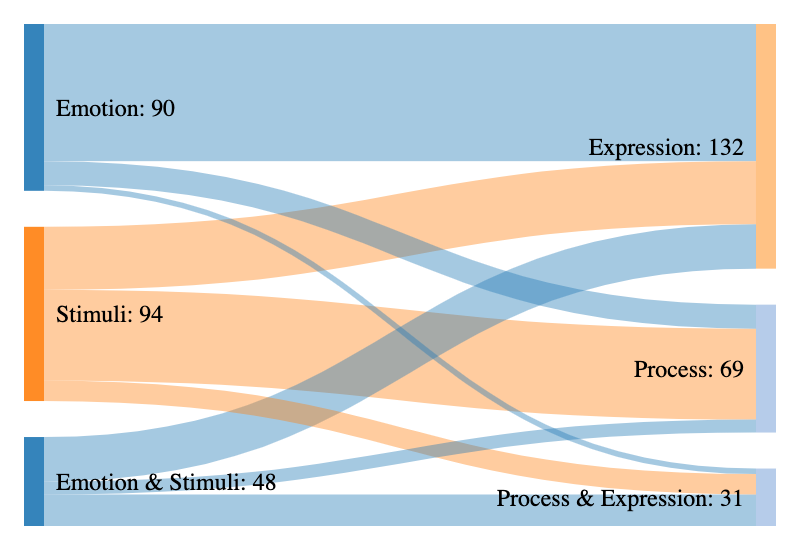}
  \caption{Sankey Diagram of Input to Output}
  \label{fig:sankey}
\end{figure}
\section{Results}

\begin{table*}[]
\centering
\begin{tabular}{@{}lcccccc@{}}
\toprule
                  & \multicolumn{1}{l}{1990 - 2000 (n=16)} & \multicolumn{1}{l}{2001-2005 (n=29)} & \multicolumn{1}{l}{2006-2010  (n=61)} & \multicolumn{1}{l}{2011-2015  (n=72 )} & \multicolumn{1}{l}{2016-2019  (n=52 )} & \multicolumn{1}{l}{Total} \\ \midrule
History           & 6.67\%                                 & 27.59\%                              & 32.79\%                               & 22.22\%                                & 34.62\%                                & 27.16\%                   \\ \midrule
Emotion Core      & 68.75\%                                & 31.03\%                              & 72.13\%                               & 70.83\%                                & 69.23\%                                & 65.95\%                   \\
Emotion Component & 31.25\%                                & 68.97\%                              & 27.87\%                               & 29.17\%                                & 30.77\%                                & 34.05\%                   \\ \midrule
Interaction       & 62.50\%                                & 65.52\%                              & 80.33\%                               & 76.39\%                                & 80.77\%                                & 75.86\%                   \\
Performance       & 37.50\%                                & 34.48\%                              & 19.67\%                               & 23.61\%                                & 19.23\%                                & 24.14\%                   \\ \midrule
Digital           & 37.50\%                                & 44.83\%                              & 39.34\%                               & 31.94\%                                & 50.00\%                                & 40.09\%                   \\
Robot             & 62.50\%                                & 55.17\%                              & 60.66\%                               & 68.06\%                                & 50.00\%                                & 59.91\%                   \\ \bottomrule
\end{tabular}
  \caption{Percentage Breakdown of History, the Role, Purpose and Implementation}
  \label{tab:bigtable}
\end{table*}

\subsection{Broad Categories}
An analysis of the results from the categorization of abstracts shows a continual growth in the research of robotics and emotion across all categories. Figure \ref{fig:broad} shows these trends for each category as well as the specific growth in input categorization. While we categorized input, we did not create clear categories for the output. We found that output was commonly multi-modal and often focused on the robot as a whole, without  always a clear emphasis. Our categorization for input includes movement as separate category, designed for cases where the emphasis is on a particular aspect of continuous movement such as gait. The clear spike in the the graphs for input corresponds to the growth in facial emotion recognition. This increase in robotics research and face recognition mirrors the recent leap in deep learning and face recognition \cite{masi2018deep}, beginning from the state of the art paper DeepFace in 2014 \cite{taigman2014deepface}.

\subsection{Emotional Intelligence - Algorithm}
The primary algorithm used within our classification system was computational complex. Figure \ref{fig:alogrithmusecoun} shows the spread of algorithm use. We did not find any unifying trends amongst algorithm use. Neural networks are the main outlier with 11 of their 12 uses happening after 2012, with continual growth through this time period.

\subsection{Emotional Intelligence - History}
The total use of history in a system was 27.16\% with limited trends over time. From 1990-2000 only one paper from the sixteen we analyzed included history, however all following time periods fit in the range of 27\% to 35\% as shown in Table \ref{tab:bigtable}.

\subsection{Emotional Intelligence - Mapping}
In the categorization of input to output, all possible combinations were represented in at least 3 publications in the dataset, shown in Figure \ref{fig:sankey} and Table \ref{tab:inputcat}. Table \ref{tab:inputcat} shows for performance or interaction which types of input were used and the frequency; in the first row it shows performance systems with emotional stimuli mapped to internal process occurs five times in the dataset. Emotion to Process \& Expression, and Emotion \& Stimuli to process had the lowest use, in 3 and 7 papers respectively. In contrast Emotion to Expression was by far the most common mapping being used 74 times, while Stimuli to Process was the second most common with 49 uses. 

\begin{table}[h]
\centering
\begin{tabular}{@{}llllll@{}}
\toprule
            & Stimuli & Emotion & Expression & Process & Freq \\ \midrule
Performance & No      & Yes     & No         & Yes     & 5    \\
Performance & Yes     & No      & No         & Yes     & 40   \\
Performance & Yes     & No      & Yes        & No      & 2    \\
Performance & Yes     & No      & Yes        & Yes     & 3    \\
Performance & Yes     & Yes     & No         & Yes     & 4    \\
Performance & Yes     & Yes     & Yes        & Yes     & 2    \\
Interaction & No      & Yes     & No         & Yes     & 8    \\
Interaction & No      & Yes     & Yes        & No      & 74   \\
Interaction & No      & Yes     & Yes        & Yes     & 3    \\
Interaction & Yes     & No      & No         & Yes     & 9    \\
Interaction & Yes     & No      & Yes        & No      & 32   \\
Interaction & Yes     & No      & Yes        & Yes     & 8    \\
Interaction & Yes     & Yes     & No         & Yes     & 3    \\
Interaction & Yes     & Yes     & Yes        & No      & 24   \\
Interaction & Yes     & Yes     & Yes        & Yes     & 15   \\ \bottomrule
\end{tabular}
  \caption{Input and Output Mapping By System Purpose}
  \label{tab:inputcat}
\end{table}

\subsection{Emotional Model - Model and Categories}
Custom emotional models (n = 154) were used in 67\% of the papers analyzed. The discrete Ekman emotions (n = 45) were the second most common occurring in 19\% of the publications. Circumplex (n = 20) , PAD (n =8 ), and Plutchik (n = 5) each occurred in less than 10\% of the puplications. 

\begin{figure}[h]
  \centering
\includegraphics[width=8cm]{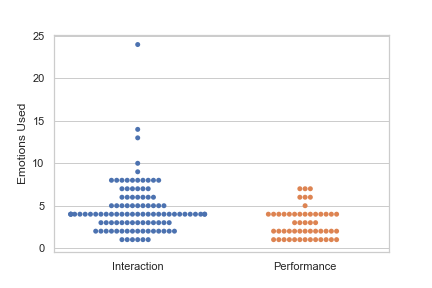}
  \caption{Quantity of Emotions Used in Custom Emotion Models}
  \label{fig:emotionsused}
\end{figure}

Figure \ref{fig:emotionsused} shows the number of emotion categories used by all publications with custom emotion models; a paper that uses happy and sad would show two emotions used. The mean for all categories was 4.10 with a median of 4. Publications with a purpose of interaction had a mean of 4.66 and median 4, while performance publications had a mean of 2.97 and median of 2.

\subsection{Implementation - Role, Purpose and Platform}
The role of emotion as either the core research or as a component of research has varied only slightly over time. Outside of a jump between 2001-2005 the range has remained close to the current total of 66\% as the core, and 34\% as a component. The purpose of emotion - either for interaction or performance - has seen a gradual shift towards an emphasis on use in interaction. From 2016-2019, 81\% of emotion related papers used emotion for interactions. Across all publications interaction was the focus of 76\% of papers. 

Table \ref{tab:bigtable} shows the variation between robot and digital implementations. There were no significant trends with physical robots being used slightly more. The most commonly used robot was SoftBank Robotics Nao with 14 uses. The other robots used were a mix of commercial and research robots like Pepper \cite{kashii2017ex} and many custom designs. The mean usage of each robot platform was 1.44 times with a median of a single use. As expected when a custom designed robot was used multiple times it was exclusively by the same researchers.



\section{Discussion}
From the development of our method and results we found multiple trends that require further analysis and discussion. In the following section we discuss both clear findings from the publications as well as our own beliefs on possible future directions and areas that require more refinement. These points are divided into three sections, paradigm features, future opportunities and intrinsic challenges. Paradigm features are clear trends that are invoked across all publications analyzed. Future opportunities and intrinsic challenges describe our subjective views for future work and areas we believe have room to develop within current literature.

\subsection{Paradigm Features}
\subsubsection{Distinct Approaches for Interaction and Performance}
While not unexpected, the results clearly present two methodologies associated with the role of emotion in a robotic system for interaction or performance. For interaction and performance the emotional models and number of categories used varies greatly. Likewise the mapping system used consistently falls into different categories. While this distinction was expected it is clear that robotics and emotion studies should be compared with a framework to their related purpose.

\subsubsection{Diverse Approaches Shaped by Current Trends}
Both the algorithms used and the emotional models used showed a wide diversity of approaches. The computational basic and complex categories, which featured custom models not clearly categorized, represented over 60\% of the approaches. While there is wide variation the literature does follow broader computer science and engineering trends. This is most clear in the significant increase in facial analysis for emotion. Likewise, 11 out of the 12 neural networks used in emotional models were used since 2012, and it is reasonable to expect more work in this area, shadowing neural networks overall growth across many domains.

\subsection{Future Opportunities}
\subsubsection{Limited Long Term Interaction and History}
Our definition of a model containing history allowed for the lowest inclusion, such as a single previous step or a first order Markov model. Even within this context only a total of 27\% of papers included history within their model. In the majority of publications history was rarely considered in more than one or two previous steps. Beyond adding short-term emotion to more systems there is clearly a wider-scope for systems that have longer term emotions, carry across the entirety of interactions and even emotional models that carry day-to-day developments within the robot. 

\subsubsection{Signalling is the Current Paradigm for Interaction}
Our analysis of papers reconfirmed ideas presented in other literature of signalling being the dominant paradigm in human-robot interaction with emotions. Emotional signaling relies on the idea that emotions are expressed and reveal the inner state of the robot. Signaling has many limitations, such as an assumption that an outwardly shown state is always an internal emotion. It also implies that displaying an emotion always carries the same meaning to the other participants \cite{jung2017affective}. Signalling further implies that all parts of the emotion are displayed, when often a human may only display part of the emotion they are feeling \cite{bucci2019real}. In our review we found signaling was overlooked unanimously by each publication.

\subsubsection{The Social Aspect of Emotions}
In psychology research, emotions are considered inherently social \cite{van2016social}, with emotions shaped by our interactions with others. This occurs through direct influence from others, as well as regulation based on social expectations \cite{van2009emotions,parkinson1996emotions}. In most papers, emotion was seen as at most a dyadic occurrence and often as a solitary experience held only by the robot. The omission of social aspects prevents many psychology based implementations and risks missing a crucial aspect of human emotional models. Some promising results for analyzing interactions between teams of humans and robots has been shown \cite{correia2018group}, but social based emotions as a whole have very limited representation in the publications analyzed. 

\subsection{Intrinsic Challenges}

\subsubsection{Anthropomorphism}
While research into robotics and emotion will inherently cross into anthropomorphism, we believe extra consideration should be given to the language used in describing robotic emotion systems. It was common in papers for a robot to be described as 'feeling' when it had a single input mapped to an emotion. Implications that a robot actually feels sad when a simple negative event happens overly reduces and simplifies much of the research done in artificial emotions. 


\subsubsection{Custom Emotional Models}
There are many theories and models of emotion published in psychology research, with a wide variety of reviewed and generally accepted models. While custom models of emotion are at times certainly appropriate, we believe many papers slightly varied established models such as Ekman's without providing a clear rationale behind the variation. This variation without proper justification discourages proper evaluation between systems. A related survey in affective computing also found that the majority of papers used custom emotional models, demonstrating this is not only a feature of robotic systems \cite{sreeja2017emotion}.

\subsubsection{Project Isolation}
A significant challenge of analyzing papers on robotics and emotion is the relative isolation of each system. As previously stated, even fundamental aspects such as the emotional model used varied greatly between each project. In addition to challenges in comparison between publications, it is often not possible to compare each system to anything outside of the proposed control from within the study. In our review we found only one paper that compared their emotional model with a baseline system \cite{godfrey2009towards}, every other publication compared only one emotional model usually to a control group which had no emotion added.

While some papers compare a digital implementation with a physical implementation, it was very rare for a paper to compare an emotional model on more than one robot. We found a single paper\cite{park2009} that compared the same model on different physical implementations. While we acknowledge testing implementations on different robots adds significant scope, we believe there is room for much more work in this area, especially considering the nature of emotion. This includes ideas such as emotional models having different applications and implications on humanoid and non-humanoid robots. Currently we can only infer from different publications, but lack controlled comparisons between models. This issue is further compounded with consideration that the average use of each robot in the publications was 1.44 times for an emotional study, indicating there is an absence of long-term studies of emotion on the vast majority of platforms.

\section{Conclusion}
In this paper we have presented a meta-analysis of publications focused on robotics and emotion. From this work we have identified multiple trends and developed a variety of discussion points for future work in robotics and emotion. From these trends we have described features of the paradigm of emotion in robotics. With the current growth in research we believe there are still key areas that present significant challenges, primarily in the isolation of projects and variation in models used. We also believe there are many future areas that are nearly unexplored in the social role of emotions, signalling and adding history to robotic emotion.

\bibliographystyle{IEEEtran}
\bibliography{name}

\end{document}